\documentclass[10pt,twocolumn,letterpaper]{article}

\usepackage{iccv}
\usepackage{times}
\usepackage{epsfig}
\usepackage{graphicx}
\usepackage{amsmath}
\usepackage{amssymb}
\usepackage{float}
\usepackage{bbding}
\usepackage{float}

\usepackage[breaklinks=true,bookmarks=false]{hyperref}

\iccvfinalcopy 

\def\iccvPaperID{****} 

\ificcvfinal\fi

\begin{document}

\title{Deep Reinforcement Learning for Producing Furniture Layout in Indoor Scenes}

\author{Xinhan Di \Envelope $^{1}$\\
IHome Company, Nanjing\\
{\tt\small deepearthgo@gmail.com}
\and
Pengqian Yu $^{2}$\\
IBM Research,
Singapore
{\tt\small peng.qian.yu@ibm.com}
}
\maketitle
\ificcvfinal\thispagestyle{empty}\fi

\begin{abstract}
In the industrial interior design process, professional designers plan the size and position of furniture in a room to achieve a satisfactory design for selling. In this paper, we explore the interior scene design task as a Markov decision process (MDP), which is solved by deep reinforcement learning. The goal is to produce an accurate position and size of the furniture simultaneously for the indoor layout task. In particular, we first formulate the furniture layout task as a MDP problem by defining the state, action, and reward function. We then design the simulated environment and train reinforcement learning agents to produce the optimal layout for the MDP formulation. We conduct our experiments on a large-scale real-world interior layout dataset that contains industrial designs from professional designers. Our numerical results demonstrate that the proposed model yields higher-quality layouts as compared with the state-of-art model. The developed simulator and codes are available at \url{https://github.com/CODE-SUBMIT/simulator1}.
\end{abstract}

\section{Introduction}

People spend plenty of time indoors such as the bedroom, living room, office, and gym. Function, beauty, cost, and comfort are the keys to the redecoration of indoor scenes. The proprietor prefers demonstration of a layout of indoor scenes in several minutes nowadays. Many online virtual interior tools are then developed to help people design indoor spaces. These tools are faster, cheaper, and more flexible than real redecoration in real-world scenes. This fast demonstration is often based on the auto layout of indoor furniture and a good graphics engine. Machine learning researchers make use of virtual tools to train data-hungry models for the auto layout \cite{Dai_2018_CVPR,Gordon_2018_CVPR}. The models reduce the time of layout of furniture from hours to minutes and support the fast demonstration. 

Generative models of indoor scenes are valuable for the auto layout of the furniture. This problem of indoor scenes synthesis is studied since the last decade. One family of the approach is object-oriented which the objects in the space are represented explicitly \cite{10.1145/2366145.2366154,10.1145/3303766,Qi_2018_CVPR}. The other family of models is space-oriented which space is treated as a first-class entity and each point in space is occupied through the modeling \cite{10.1145/3197517.3201362}.

\begin{figure*}
\centering
\includegraphics[height=6.5cm]{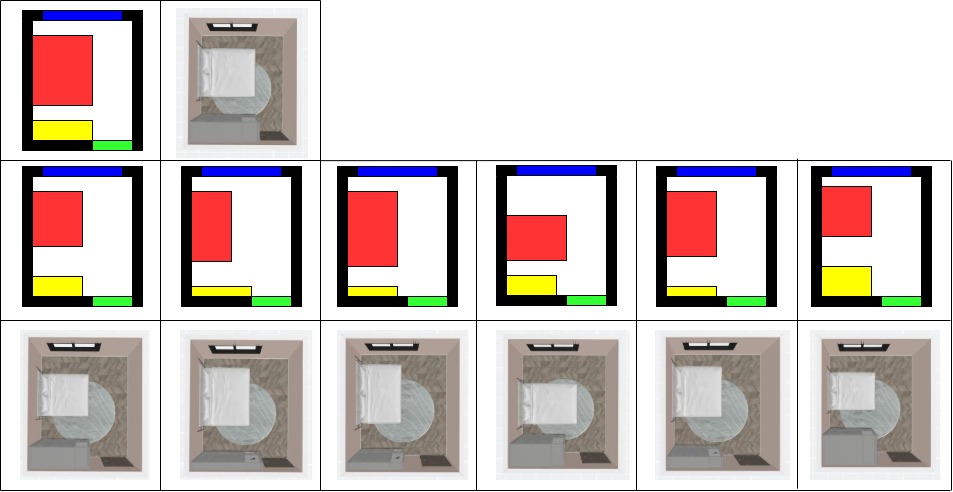}
\caption{Examples of layouts produced by the state-of-the-art models \cite{10.1145/3306346.3322941}. These layouts are for bedroom. The first row gives the ground truth layout in the simulator and the real-time renders. The second and third rows show the layouts produced by the state-of-the-art models \cite{10.1145/3306346.3322941}. From the results, the state-of-the-art model can only produce approximate position and size of the furniture.}
\label{fig1}
\end{figure*}

\begin{figure*}
\centering
\includegraphics[height=6.5cm]{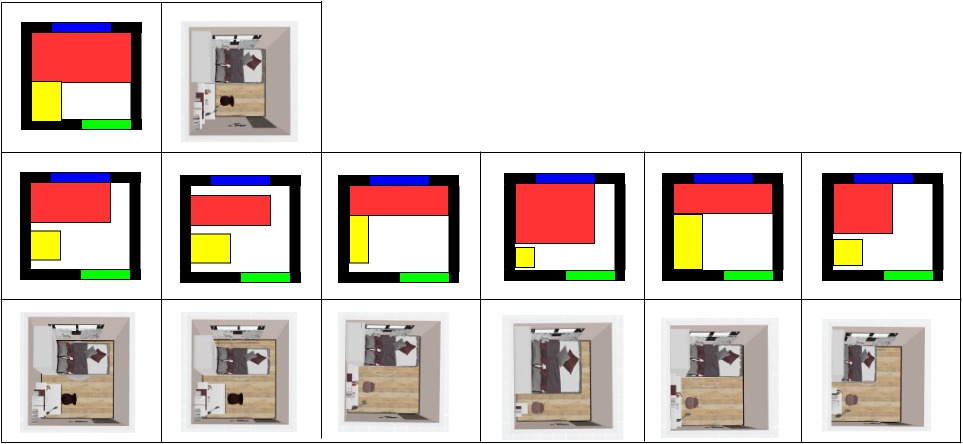}
\caption{Examples of layouts produced by the state-of-the-art models \cite{10.1145/3306346.3322941}. These layouts are for tatami room. The first row gives the ground truth layout in the simulator and the real-time renders. The second and third rows show the layouts produced by the state-of-the-art models \cite{10.1145/3306346.3322941}. From the results, the state-of-the-art model can only produce approximate position and size of the furniture.}
\label{fig2}
\end{figure*}

Deep generative models are used for efficient generation of indoor scenes for auto-layout recently. These deep models further reduce the time from minutes to seconds. The variety of the generative layout is also increased. The deep generative models directly produce the layout of the furniture given an empty room. In the real world, a satisfactory layout of the furniture requires two key factors. The first one is the correct position, and the second one is a good size. However, this family of models only provides approximate size of the furniture, which is not practical in the real world industry as illustrated in Figure \ref{fig1} and Figure \ref{fig2}.  

The prior work neglects the fact that the industrial interior design process is indeed a sequential decision-making process, where professional designers need to make multiple decisions on size and position of furniture before they can produce a high-quality of design. In practice, professional designers need to be in a real room, improve the furniture layout step by step by obtaining the feedback of the current decision until a satisfactory design is produced. This industrial process can be naturally modelled as a Markov decision process (MDP). 

Reinforcement learning (RL) consists of an agent interacting with the environment, in order to learn an optimal policy by trial and error for MDP problems. The past decade
has witnessed the tremendous success of deep reinforcement learning (RL) in the fields of gaming, robotics and recommendation systems \cite{gibney2016google,schrittwieser2020mastering,silver2017mastering}. Researchers have proposed many useful and practical algorithms such as DQN \cite{mnih2013playing} that learns an optimal policy for discrete action space, DDPG \cite{lillicrap2015continuous} and PPO \cite{schulman2017proximal} that train an agent for continuous action space, and A3C \cite{mnih2016asynchronous} designed for a large-scale computer cluster. These proposed algorithms solve stumbling blocks in the application of deep RL in the real world.   

We highlight our two main contributions to produce the layout of the furniture with accurate size. First, we develop an indoor scene simulator and formulate this task as a Markov decision process (MDP) problem. Specifically, we define the key elements of a MDP including state, action, and reward function for the problem. Second, we apply deep reinforcement learning technique to solve the MDP. This proposed method aims to support the interior designers to produce decoration solutions in the industrial process. In particular, the proposed approach can produce the layout of furniture with good position and size simultaneously.

This paper is organized as follows: the related work is discussed in Section 2. Section 3 introduces the problem formulation. The development of the indoor environment simulator and the models for deep reinforcement learning are discussed in Section 4. The experiments and comparisons with the state-of-art generative models can be found in Section 5. The paper is concluded with discussions in Section 6.

\section{Related Work}

Our work is related to data-hungry methods for synthesizing indoor scenes through the layout of furniture unconditionally or partially conditionally. 

\subsection{Structured data representation}
Representation of scenes as a graph is an elegant methodology since the layout of furniture for indoor scenes is highly structured. In the graph, semantic relationships are encoded as edges, and objects are encoded as nodes. A small dataset of annotated scene hierarchies is learned as a grammar for the prediction of hierarchical indoor scenes \cite{10.1145/3197517.3201362}. Then, the generation of scene graphs from images is applied, including using a scene graph for image retrieval \cite{Johnson_2015_CVPR} and generation of 2D images from an input scene graph \cite{Johnson_2018_CVPR}. However, the use of this family of structure representation is limited to a small dataset. In addition, it is not practical for the auto layout of furniture in the real world. 

\subsection{Indoor scene synthesis}
Early work in the scene modeling implemented kernels and graph walks to retrieve objects from a database \cite{Choi_2013_CVPR,Dasgupta_2016_CVPR}. The graphical models are employed to model the compatibility between furniture and input sketches of scenes \cite{10.1145/2461912.2461968}. However, these early methods are mostly limited by the size of the scene. It is therefore hard to produce a good-quality layout for large scene size. With the availability of large scene datasets including SUNCG \cite{Song_2017_CVPR}, more sophisticated learning methods are proposed as we review them below.

\subsection{Image CNN networks}
An image-based CNN network is proposed to encoded top-down views of input scenes, and then the encoded scenes are decoded for the prediction of object category and location \cite{10.1145/3197517.3201362}. A variational auto-encoder is applied to the generation of scenes with the representation of a matrix. In the matrix, each column is represented as an object with location and geometry attributes \cite{10.1145/3381866}. A semantically-enriched image-based representation is learned from the top-down views of the indoor scenes, and convolutional object placement a prior is trained \cite{10.1145/3197517.3201362}. However, this family of image CNN networks can not apply to the situation where the layout is different with different dimensional sizes for a single type of room.

\subsection{Graph generative networks}
A significant number of methods have been proposed to model graphs as networks \cite{DBLP:journals/corr/abs-1709-05584,4700287}, the family for the representation of indoor scenes in the form of tree-structured scene graphs is studied. For example, Grains \cite{10.1145/3303766} consists of a recursive auto-encoder network for the graph generation and it is targeted to produce different relationships including surrounding and supporting. Similarly, a graph neural network is proposed for the scene synthesis. The edges are represented as spatial and semantic relationships of objects \cite{10.1145/3197517.3201362} in a dense graph. Both relationship graphs and instantiation are generated for the design of indoor scenes. The relationship graph helps to find symbolical objects and the high-lever pattern \cite{10.1145/3306346.3322941}. However, this family of models hard produces accurate size and position for the funiture layout. 

\subsection{CNN generative networks}
The layout of indoor scenes is also explored as the problem of the generation of the layout. Geometric relations of different types of 2D elements of indoor scenes are modeled through the synthesis of layouts. This synthesis is trained through an adversarial network with self-attention modules \cite{DBLP:journals/corr/abs-1901-06767}. A variational autoencoder is proposed for the generation of stochastic scene layouts with a prior of a label for each scene \cite{Jyothi_2019_ICCV}. However, the generation of the layout is limited to produce a similar layout for a single type of room. 

\section{Problem Formulation}
We formulate the process of furniture layout in the indoor scene as a Markov Decision Process (MDP) augmented with a goal state $G$ that we would like an agent to learn. We define this MDP as a tuple $(S,G,A,T,\gamma)$, in which $S$ is the set of states, $G$ is the goal, $A$ is the set of actions, $T$ is the transition probability function in which $T(s,a,s^{'})$ is the probability of transitioning to state $s^{'}$ when action $a$ is taken in state $s$, $R_t$ is the reward function at timestamp $t$, $\gamma$ is the discount rate $\gamma\in[0,1)$. At the beginning of each episode in a MDP, the solution to a MDP is a control policy $\pi: S,G \rightarrow A$ that maximizes the value function $v_{\pi}(s,g):=\mathbb{E}_{\pi}[\sum_{t=0}^{\infty} \gamma^{t} R_{t}|s_{0}=s,g=G]$ for given initial state $s_0$ and goal $g$.

\begin{figure*}
\centering
\includegraphics[height=8.0cm]{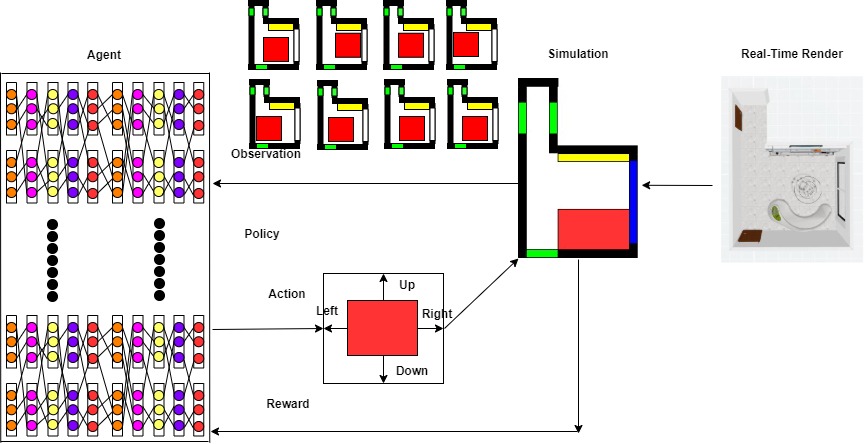}
\caption{The Formulation of MDP for the layout of furniture in the indoor scenes. A simulator is developed to bring the indoor scene in the real-time render into the simulation. Then, deep reinforcement learning is applied to train an agent for the exploration of action, reward, state and optimal policy for the developed simulated environment.}
\label{fig3}
\end{figure*}

As shown in Figure \ref{fig4}, the states $S$ is a tuple $S=(S_{ws},S_{wins},S_{fs})$ where $S_{ws}$ is the geometrical position $p$, size $s$ of the wall $ws$, To be noted, $p$ is the position $(x,y)$ of the center of an wall, size $s$ is the width and height of the wall. Similarly, the geometrical position $p$, size $s$, of the windows $wins$ and doors $ds$ are defined as the position of the center and the (width,height) of the wall, and the geometrical position $p$, size $s$ of the furniture $fs$. Therefore, the state $S$ is a tuple $S=(S_{ws},S_{wins},S_{fs})$, where $S_{wins}$ contains $N_{wins}$ windows with corresponding $p_{i}$, size $s_{i}$ for each windows, $i \in \{1,2,\dot,N_{wins}\}$. Similarly, $S_{ws}$ contains $N_{ws}$ walls with corresponding $p_{i}$, size $s_{i}$ for each wall, $i \in \{1,2,\dot,N_{ws}\}$. $S_{fs}$ contains $N_{fs}$ furniture with corresponding $p_{i}$, size $s_{i}$ for each furniture, $i \in \{1,2,\dot,N_{fs}\}$. The goal $G$ is represented as the correct position $p$, size $s$ and direction $d$ of the furniture. The correct value of the furniture is from the design of professional interior designer which is sold to the customer. The action $A$ represents the motion of how the furniture moves to the correct position $p$.  

\begin{figure*}
\centering
\includegraphics[height=6.5cm]{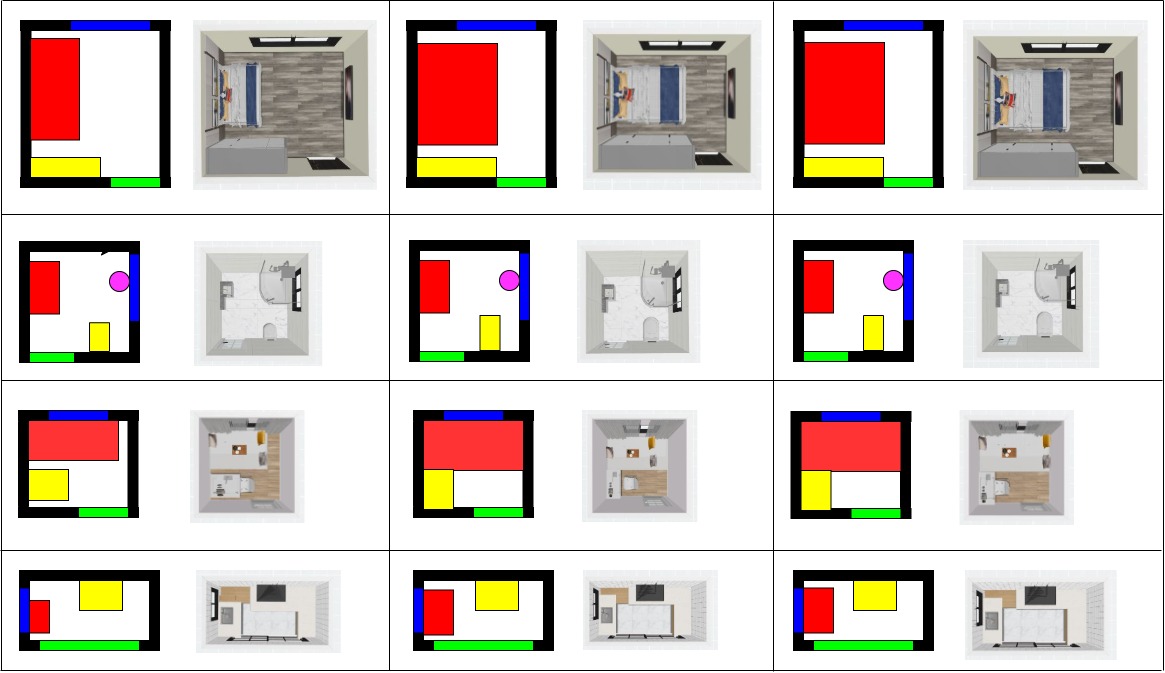}
\caption{Evaluation sample of the furniture layout for the bedroom, the bathroom, the tatami room and the kitchen. The first column represents the results of the state-of-the-art models \cite{10.1145/3306346.3322941}. The second column represents the results of the proposed method in the simulation. The third column represents the ground truth layout. The state-of-the-art model is able to predict the approximate size and position of the furniture, while the proposed method is able to predict more accurate size and position. The prediction is close to the ground truth.}
\label{fig6}
\end{figure*}

\section{Methods}
\begin{figure*}
\centering
\includegraphics[height=6.5cm]{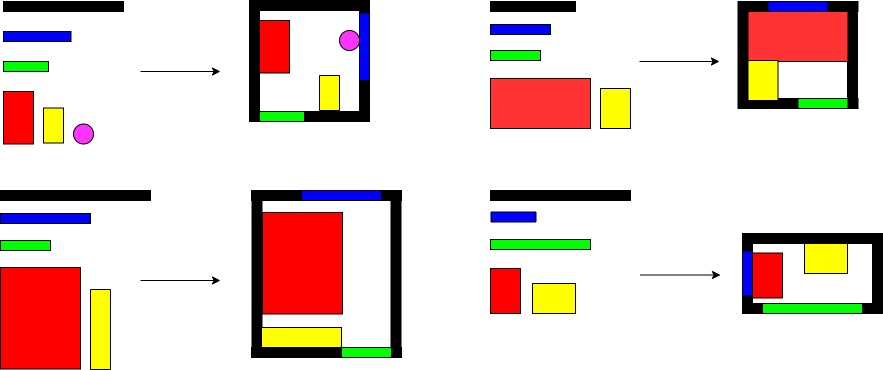}
\caption{The simulator transfer the indoor scenes into a simulated indoor scenes. The walls are represented as black patterns. The doors are represented as green patterns. The windows are represented as blue patterns. For the bathroom, shower is represented as the pink pattern, the toilet is represented as the yellow pattern, the bathroom cabinet is represented as the red pattern. For the bedroom, the bed is represented as the red pattern, the cabinet is represented as the yellow pattern. For the tatami room, the tatami is represented as the red pattern, the working desk is represented as the yellow pattern. For the kitchen, the sink is represented as the red pattern, the cook-top is represented as the yellow pattern.}
\label{fig4}
\end{figure*}

\begin{figure}
\includegraphics[height=4.0cm]{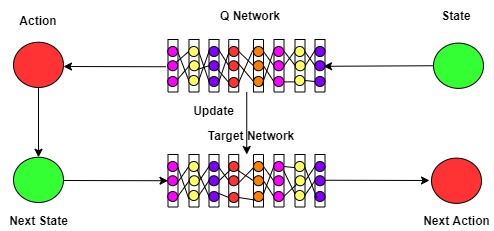}
\caption{DQN network architecture combined with the developed simulation in the formulated MDP. }
\label{fig5}
\end{figure}
In order to solve this formulated MDP problem. We explore and develop the simulated environment, action, reward, agent and the training of the agent in this section as Figure \ref{fig3} and Figure \ref{fig4} shown. Briefly, we define all elements in the indoor scene for this formulated MDP problem. Then, a real-time simulator is developed to explore this dynamics. In order to solve this MDP problem, we explore the reward, action, agent and policy in the filed of the reinforcement learning. Finally, a reinforcement learning model is trained to find a optimal policy as a solution to this formulated MDP problem.     
\begin{figure*}
\centering
\includegraphics[height=6.5cm]{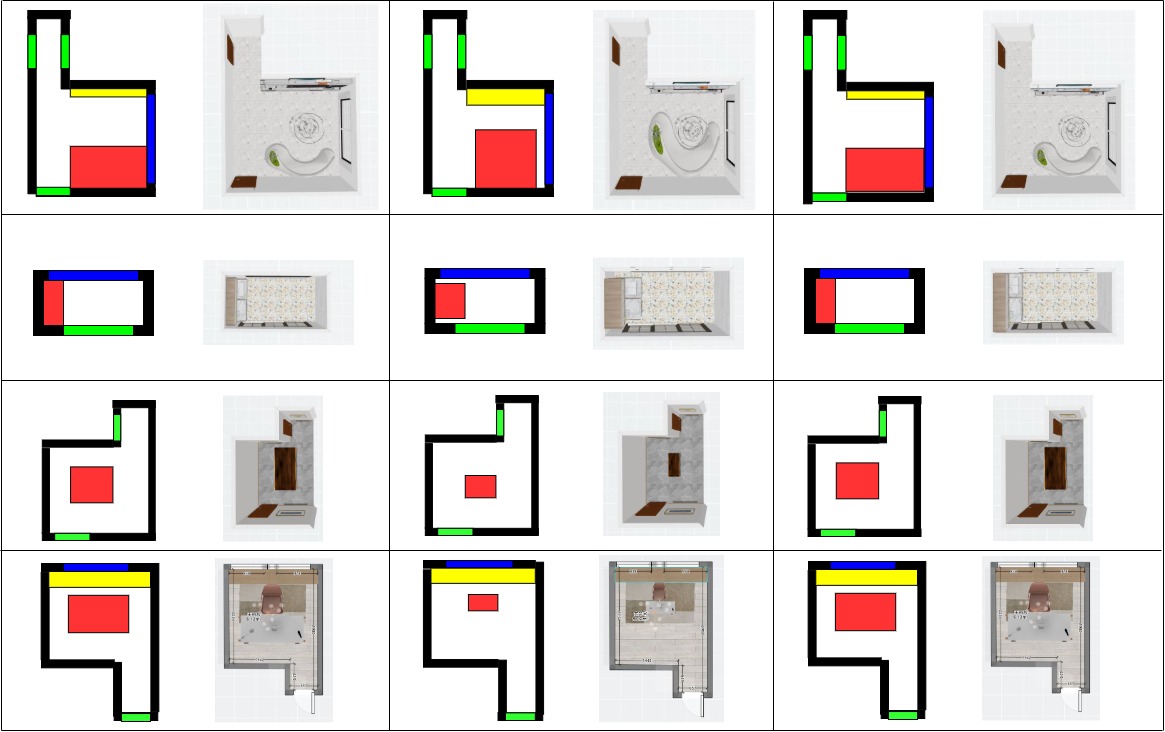}
\caption{Evaluation sample of the furniture layout for the living-room, the balcony, the dining-room room and the study room. The first column represents the results of the state-of-the-art models \cite{10.1145/3306346.3322941}. The second column represents the results of the proposed method in the simulation. The third column represents the ground truth layout. The state-of-the-art model is able to predict the approximate size and position of the furniture, while the proposed method is able to predict more accurate size and position. The prediction is close to the ground truth.}
\label{fig7}
\end{figure*}

\subsection{Environment}
This indoor environment for this defined MDP is implemented as a simulator $F$, where $(S_{next},R)=F(S,A)$, $A$ is the action from the agent in the state $S$. The simulator $F$ receives the action $A$ and produces the next state $S_{next}$ and reward $R$. The simulator builds a world of a room contains the walls, windows, doors and the furniture. As Figure \ref{fig4} shown, the walls are visualized as the black patterns, the windows are visualized as the blue patterns, the doors are visualized as the green patterns. The furniture are visualized as different color of the patterns. For example, the cabinet is visualized as yellow patterns, the bed is visualized as red patterns. The simulator receives action and then  produces the reward and the next state. In the next state, the geometrical position, size of walls, doors, and windows are not changed, the geometrical position of the furniture is updated following the input action.

\begin{figure*}
\centering
\includegraphics[height=6.5cm]{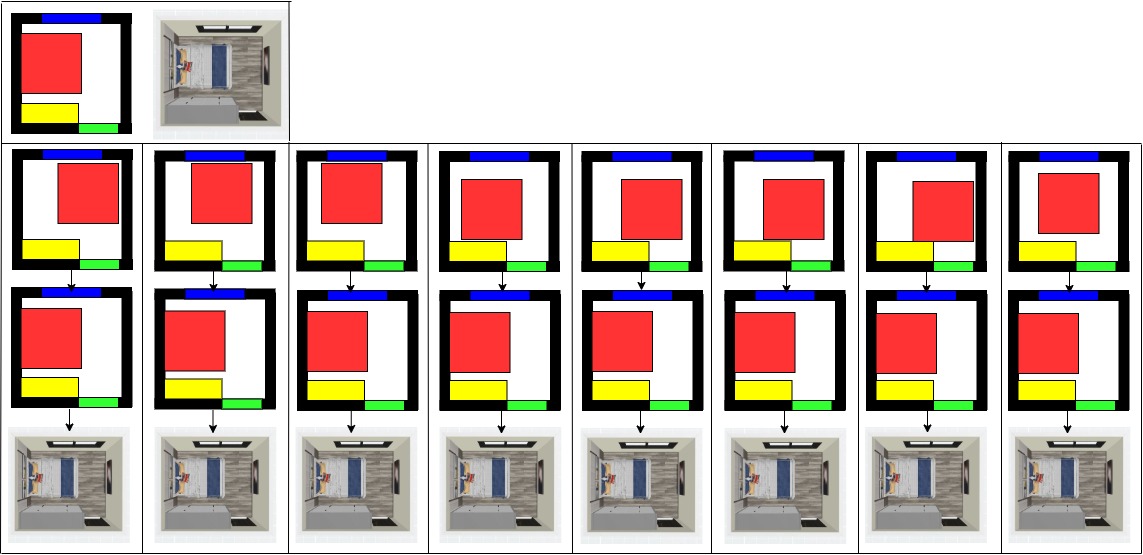}
\caption{Given a bed with random positions, the proposed method is able to produce a good layout of the bed in the bedroom. The first row represents the the ground truth layout for a bedroom in the simulation and corresponding renders. The second row represents the bed in random positions. The third row represents the final layout produced by the proposed method. The fourth row represents the corresponding layout in the renders.}

\label{fig8}
\end{figure*}

\subsection{Action}
The action space is discrete, they are defined as four different action named as right, left, below and up. It means the center of the furniture moves right, left, below and up in a step. Besides, we set another two rules particularly for this simulator. Firstly, in the training process, if the furniture moves beyond the room, the simulator drops this action and then receives a new action. In the test process, if the furniture moves beyond the room, the simulator stops the action immediately. As Figure \ref{fig3} shown.

\begin{figure*}
\centering
\includegraphics[height=6.5cm]{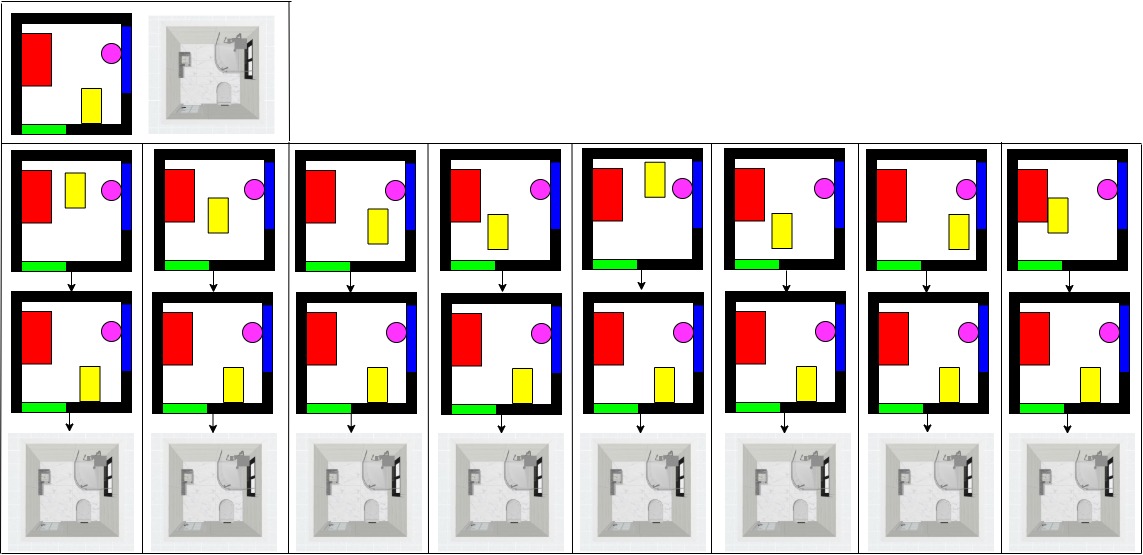}
\caption{Given a toilet with random positions, the proposed method is able to produce a good layout of the toilet in the bathroom. The first row represents the the ground truth layout for a toilet in the simulation and corresponding renders. The second row represents the toilet in random positions. The third row represents the final layout produced by the proposed method. The fourth row represents the corresponding layout in the renders.}

\label{fig9}
\end{figure*}

\subsection{Reward}
We define three types of reward for this environment. Firstly, the first reward function works to encourage the furniture to move towards the most right position. It's defined as the following:

\begin{equation}
    r1=\theta_{1}IoU(f_{target},f_{state})
\end{equation}

where $\theta_{1}$ is the parameter which is a positive parameter, $f_{target}$ represents the ground truth position and size of the furniture, $f_{state}$ represents the current position and size of the furniture in the state. IoU represents the intersection between $f_{state}$ and $f_{target}$.

The second reward function works to prevent the furniture moving in coincide with other elements including the walls, windows, doors and other furniture. It's defined as the following:

\begin{multline*}
    r2=\theta_{2}(\sum_{i=0}^{N_{fs}-1}IoU(f_{state},f^{'}_{target})+\sum_{i=0}^{N_{ws}}IoU(f_{state}, w_{target})\\
    +\sum_{i=0}^{N_{ds}}IoU(f_{state},d_{target})+\sum_{i=0}^{N_{wins}}IoU(f_{state},wins_{target}))
\end{multline*}

where $\theta_{2}$ is the parameter which is a negative parameter, $f_{state}$ represents the current position and size of the furniture in the state, $f^{'}_{target}$ represents the ground truth position and size of other furniture in the environment, $w_{target}$ represents the ground truth position and size of the walls in the environment, $d_{target}$ represents the ground truth position and size of the doors in the environment, $wins_{target}$ represents the ground truth position and size of the windows in the environment.
IoU represents the intersection between the ground truth object and the predicted object.

The third reward function works to prevent the furniture moving outside the room. It's defined as the following:

\begin{equation}
    r3=\theta_{3}(1-IoU(f_{state},Indoor_{area})) 
\end{equation}

where $\theta_{3}$ is the parameter which is a negative parameter,  $f_{state}$ represents the current position and size of the furniture in the state for the furniture. $Indoor_{area}$ represents the area of the indoor rooms. IoU represents the intersection between the predicted object and the area of the indoor room.

Then, the mutual reward is calculated as the sum of the three type reward, $r=\sum_{i=1}^{3}r_{i}$. It's aimed at moving the furniture to the right position with suitable size, preventing the furniture moving in coincide with walls, doors, windows and other furniture. Besides. it's aimed at moving the furniture outside the room.

\subsection{Agent}
Consider the above Markov decision process defined by the tuple $\mu=(S,G,A,T,\gamma)$ where the state $S$, goal $G$, action $A$ is defined as above. And the reward $R$ is defined as above. Commonly, for an agent, upon taking any action $a \in A$ at any state $s \in S$, $P(·|s, a)$ defines the probability distribution of the next state and $R(·|s, a)$ is the distribution of the immediate reward. In practise, the state for this environment is a compact subset $S={S_{1},S_{2},\dot,S_{n}}$, where each the walls, windows, doors and other furniture in each state $S_{i},i\in \{1,\dot,n\}$ keeps non-changed, only the furniture moving by the agent is changed to different positions and sizes. $A={a_{1},a_{2},a_{3},a_{4}}$ has finite cardinality $4$, $a_{1}$ represents moving left, $a_{2}$ represents moving right, $a_{3}$ represents moving up, $a_{4}$ represents moving below. The rewards $R(·|s, a)$ has a range on $[-200,100]$ for any $s \in S$ and $a \in A$.     

The agent learns a policy $\pi:S \rightarrow P(A)$ for the MDP maps any state $s \in S$ to a probability distribution $\pi(·|s)$ over $A$. For a given policy $\pi$, starting from the initial state $S_{0}=s$, the actions, rewards, and states evolve according to the law as follows:

\begin{multline*}
    (A_{t},R_{t},S_{t+1}):A_{t} \sim \pi(.|S_{t}), R_{t} \sim R(.|S_{t},A_{t}),\\
    S_{t+1} \sim P(.|S_{t},A_{t}), t=0,1,\dots,n
\end{multline*}

and the corresponding value function $V_{\pi}:S \rightarrow R$ is defined as the cumulative discounted reward obtained by taking the actions according to $\pi$ when starting from a fixed state , that is,

\begin{equation}
    V^{\pi}(s) = \mathcal{E}[\sum_{t=0}^{\infty}\gamma^{t}·R_{t}|S_{0}=s]
\end{equation}

The policy $\pi$ is controlled by the agent, the functions $P$ and $R$ are defined above for the environment. By the law of iterative expectation, for nay policy $\pi$.

\begin{equation}
    V^{\pi}(s) = \mathcal{E}[Q^{\pi}(s,A)|A \sim \pi(.|s)], \forall s \in S
\end{equation}

where $Q^{\pi}(s,a)$ is the action value function. The agent is trained to find the optimal policy, which achieves the largest cumulative reward via dynamically learning from the acquired data. The optimal action-value function $Q^{*}$ is defined as following:

\begin{equation}
    Q^{*}(s,a) = \sup_{\pi}Q^{\pi}(s,a), \forall (s,a) \in S \times A 
\end{equation}

where the supremum is taken over all policies. 

Then we train the agent to get the optical action value function applying a deep neural network $Q_{\theta}:S \times A \rightarrow \mathcal{R}$ is trained to approximate $Q^{*}$, where $\theta$ is the parameter as Figure \ref{fig5} shown. Simultaneously, two tricks are applied in the training of the DQN networks.

Firstly, the experience replay is applied at each time $t$, the transition $(S_{t}, A_{t}, R_{t}, S_{t+1})$ is stored into replay memory $\mathcal{M}$. Then, a mini-batch of independent samples from $M$ is sampled to train neural network via stochastic gradient descent. The goal of experience replay is to obtain uncorrelated samples, which yield accurate gradient estimation for the stochastic optimization problem. However, to be noted, the state where the moving furniture is outside the room is not stored in the memory $M$.

Secondly, a target network $Q_{\theta^{*}}$ with parameter $\theta^{*}$ is applied in the training. It uses independent samples ${(s_{i},a_{i},r_{i},s_{i+1})}i\in[n]$ from the replay memory. The parameter $\theta$ of the Q-network is updated once every $T_{target}$ steps by letting $\theta_{*}=\theta$. That is, the target network is hold fixed for $T_{target}$ steps and then it's updated by the current weights of the Q-network.

\begin{table}[t]
\centering
\begin{tabular}{|p{2cm}|p{2cm}|p{2cm}|p{2cm}|}
\hline
\multicolumn{1}{|c|}{}&\multicolumn{2}{|c|}{\text{IoU}}\\
\hline
\hfil Model    &\hfil PlanIT & \hfil Ours\\
\hline
\hfil Bathroom    &$0.623\pm0.008$ &$0.961\pm 0.014$\\
\hfil Bedroom     &$0.647\pm0.009$ &$0.952\pm 0.026$\\
\hfil Study       &$0.619\pm0.006$ &$0.957\pm 0.018$ \\
\hfil Tatami      &$0.637\pm0.006$ &$0.948\pm 0.049$ \\
\hfil Living-Room &$0.603\pm0.006$ &$0.953\pm 0.037$ \\
\hfil Dining-Room &$0.611\pm0.006$ &$0.949\pm 0.029$ \\
\hfil Kitchen     &$0.636\pm0.006$ &$0.953\pm 0.041$ \\
\hfil Balcony     &$0.651\pm0.006$ &$0.951\pm 0.035$ \\
\hline
\end{tabular}
\caption{Comparison with the state-of-art Model.}
\end{table}\label{table1}

\begin{table}[t]
\centering
\begin{tabular}{|p{2cm}|p{2cm}|p{2cm}|}
\hline
\multicolumn{1}{|c|}{}&\multicolumn{1}{|c|}{\text{IoU}}\\
\hline
\hfil Model    & \hfil Ours\\
\hline
\hfil Bathroom    &$0.953\pm0.012$ \\
\hfil Bedroom     &$0.959\pm0.017$ \\
\hfil Study       &$0.946\pm0.014$ \\
\hfil Tatami      &$0.949\pm0.009$ \\
\hfil Living-Room &$0.956\pm0.018$ \\
\hfil Dining-Room &$0.962\pm0.005$ \\
\hfil Kitchen     &$0.957\pm0.015$ \\
\hfil Balcony     &$0.961\pm0.007$ \\
\hline
\end{tabular}
\caption{Evaluation with initial random positions.}
\end{table}\label{table1}

Both the Q network and the target network contain three convolution layers and three fc layers. The convolution layers contain $8$, $16$, $32$ features, the fc layers contain $7200$, $512$ and $512$ features. Both the Q network and the target network are learned in the training process.

\section{Evaluation}
In this section, we present qualitative and quantitative results demonstrating the utility of our proposed model and developed simulation environment. Eight main types of indoor rooms are evaluated including the bedroom, the bathroom, the study room, the kitchen, the tatami room, the dining room, the living room and the balcony. We compared the proposed model with the state-of-art models, Besides, for each room, we also test the performance of the proposed model in the developed environment with $2000$ random starting points. For the comparison, we train $5k$ samples for each type of rooms and test $1k$ samples for the corresponding type of rooms. All of the samples are from the designs of professional designers.   

\subsection{Evaluation metrics}
For the task of interior scene synthesis, we apply a metric for the evaluation. For the comparison between the proposed method with the state-of-the-art models, the average IoU is calculated for each type of rooms. The IoU is defined as the following:

\begin{equation*}
    IoU_{average} = \frac{\sum_{0}^{n} IoU(f_{pred},f_{gt})}{n}
\end{equation*}

where $f_{gt}$ represents the ground truth position and size of the furniture, $f_{pred}$ represents the predicted position and size of the corresponding furniture. IoU represents the intersection between the ground truth object and the predicted object.

We compare the state-of-the-art models for scene synthesis for the rooms. The results are shown in Figure \ref{fig6} to \ref{fig7}. It can be observed that our model outperforms the state-of-art models in the following two aspects. First, the predicted position of the center of the furniture is more accuracy than the state-of-the-art models. Second, the predicted size of the furniture is the same with the ground truth size. While the state-of-the-art model is only able to predict approximate size of the furniture. Similarly, Figure \ref{fig8} and Figure \ref{fig9} represents the samples for different starting points of the furniture in the indoor room environment. While the state-of-the-art model is capable to predict the position and size of the model with an empty room as input. Furthermore, the corresponding quantitative comparison and evaluation is as shown in Table \ref{table1}. 

In detail, Figure \ref{fig6} represents the predicted layout and ground truth in both the simulated environment and the rendering environment at the product end. For the evaluation sample of the bedroom.The proposed method outperforms the state-of-the-art model for the accuracy of the size and position. Similarly, Figure \ref{fig7} represents the predicted layout of the tatami room, the study room, the dining room and the living-room. The proposed method also outperforms the state-of-the-art model for the performance of the size and position.

Besides, Figure \ref{fig8} represents the evaluation samples in randomly starting positions. The proposed method is able to move the furniture to the ground truth position from random positions. For the four types of rooms including the bedroom, the bathroom, the kitchen and the balcony. Similarly, Figure \ref{fig9} represents the evaluation samples in randomly starting positions for the bathroom. However, the state-of-the-art models is not able to achieve this task that give the prediction of layout of the furniture with randomly initial positions as input.

\section{Discussion}
In this paper, we explore to step into Markov Decision Process(MDP) for deep furniture layout in indoor scenes. In addition, we study the formulation of MDP for this layout task of indoor scenes. Besides, a simulator for this layout task is developed. Then, the reward, action, policy and agent are studied under the explored MDP formulation. After this exploration to formulate this layout task to the field of the reinforcement learning, we train two models to find a optimal policy. In the En valuation, the proposed method achieves better performance than the state-of-art models on the interior layouts dataset with more accurate position, size of the furniture. 

However, there are several avenues for future work. Our method is currently limited to moving a single furniture in the simulated environment. The simulated environment is not driven though a real-time rend erring engineer, Therefore, the simulated environment is not at the lever of industrial lever.The simulator is able to support a limited number of rooms in the training process. 

{\small
\bibliographystyle{ieee_fullname}
\bibliography{egbib}
}

\end{document}